\documentclass{article}
\usepackage{spconf,amsmath,graphicx}
\usepackage{algorithm2e}

\pdfoutput=1 

\title{A Fast Learning Algorithm for Image Segmentation with Max-Pooling Convolutional Networks}
\name{Jonathan Masci$^{\star}$\sthanks{Jonathan Masci is supported by the \emph{ArcelorMittal / New AIS} project.
This work was partially supported by the \emph{Supervised Deep / Recurrent Nets} SNF grant, Project Code 140399.} \qquad Alessandro Giusti$^{\star}$ \qquad Dan Ciresan$^{\star}$ \qquad Gabriel Fricout$^{\dagger}$ \qquad J\"{u}rgen Schmidhuber$^{\star}$}
\address{$^{\star}$ IDSIA -- USI -- SUPSI, Manno -- Lugano, Switzerland \\ 
$^{\dagger}$ArcelorMittal, Maizi\`{e}res Research, Measurement and Control Dept., France}

\begin{document}
\ninept
\maketitle
\begin{abstract}
We present a fast algorithm for training MaxPooling Convolutional Networks to segment images.  This type of network yields record-breaking performance in a variety of tasks, but is normally trained on a computationally expensive patch-by-patch basis.  Our new method processes each training image \emph{in a single pass}, which is vastly more efficient.

We validate the approach in different scenarios and report a 1500-fold speed--up.  In an application to automated steel defect detection and segmentation, we obtain excellent performance with short 
training times.
\end{abstract}
\section{Introduction}
\label{sec:intro}
Image segmentation is a fundamental task for many applications ranging from medical imaging to industrial inspection systems.  Many recent segmentation approaches build on supervised machine learning techniques, and rely on a training dataset with known ground truth segmentation.

A conventional supervised segmentation pipeline is typically based on two stages operating at the level of single pixels: a) {\em feature extraction}, where each pixel is projected into a richer representation by accounting for its context; and b) {\em classification}, where class probabilities for each pixel are computed.  Once each pixel is classified, the resulting probability maps are post-processed (e.g. by enforcing smooth boundaries through filtering and thresholding, or by using techniques such as graph cuts~\cite{boykov2001fast} or level sets~\cite{tsai2003shape}). Finding the right set of features which minimizes segmentation error is a cumbersome task. The choice of features greatly affects segmentation quality.

Recent work \cite{Ciresan:2012f,farabet-pami-13,Turaga:2010,Turaga:2009} follows a different approach, using convolutional neural networks (CNN) to segment images. Here feature extraction itself is learned from data and not enforced by designers.  These approaches obtain state-of-the-art results in a very broad range of applications.

Amongst CNN variants, the MaxPooling Convolutional Network (MPCNN) has recently received a lot of attention. It obtained a long list of record-breaking results  \cite{Ciresan:2012f,masci:2012ijcnn,ciresan:2011b,ciresan:2011a}. MaxPooling layers appear fundamental for excellent performance, but their training requires to operate separately on all patches in the image. This requires a lot of computational power, a serious limitation for many industrial applications where large datasets are used and training has to be fast. 

{\bf Contribution}
We propose an efficient MPCNN training algorithm operating on entire training images, avoiding redundant computations,
making MPCNN easily applicable to huge training datasets. We validate it on the problem of steel defect detection, achieving excellent results in this important industrial application.
	

\section{Background}
\label{sec:background}
{\bf MaxPooling Convolutional Neural Networks (MPCNN)} 
are hierarchical models
alternating two basic operations, Convolution and MaxPooling.
Their key feature
 is that
they exploit the multi-dimensional structure of images via weight sharing,
learning a set of convolutional filters.
MPCNN scale well to large images
and excel in many object recognition
\cite{ciresan:2011a,ciresan:2011b,ciresan:2011c,masci:2012ijcnn} and
segmentation \cite{Ciresan:2012f,Turaga:2009,Turaga:2010} benchmarks. 
We refer to a state-of-the-art MPCNN as depicted in Figure~\ref{fig:cnn}. It
consists of several basic building blocks briefly explained here:
\begin{figure}[htbp]
\begin{center}
\includegraphics[width=1.\linewidth]{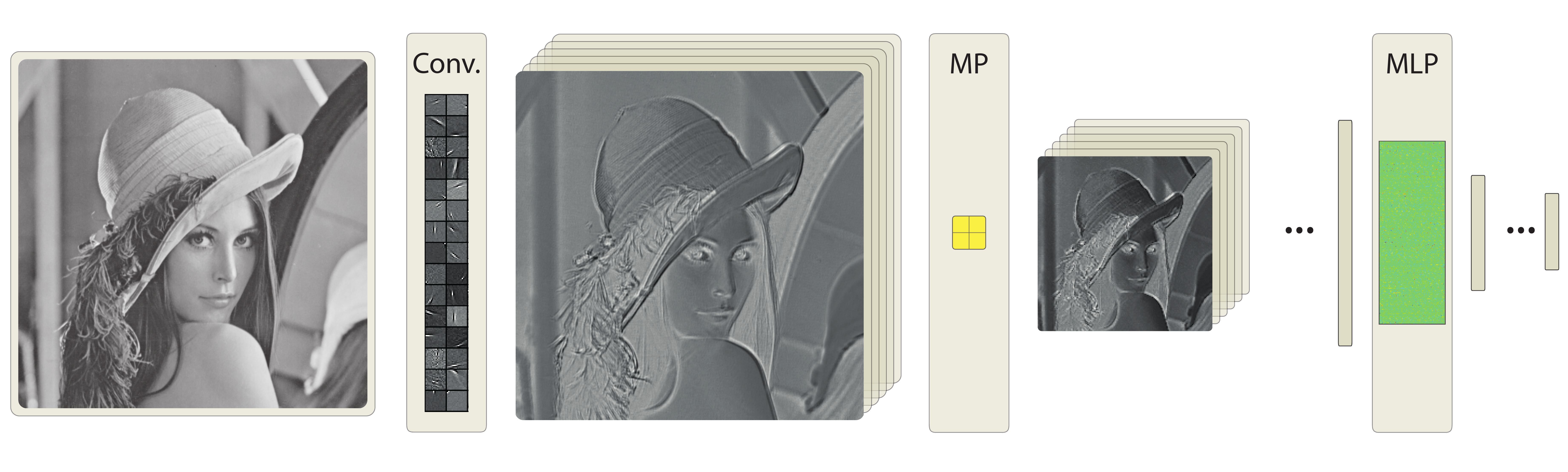}
\caption{
A schematic representation of an MPCNN.
Raw input pixel values are processed by a number of interleaved Convolutional and MaxPooling layers, which are trained to extract meaningful features.  Several Fully-Connected layers follow, which produce the final classification.
}
\label{fig:cnn}
\end{center}
\end{figure}
\\ \textit{Convolutional Layer (C)}: performs a 2D filtering between input images and a bank of filters, producing another set of images denoted as maps. Fully connected input--output correspondences are adopted and maps are linearly combined.
Then, a nonlinear activation function (e.g., $\text{tanh}$ or $\text{logistic}$) is applied.
\\ \textit{MaxPooling Layer (MP)}: down-samples the input images by a constant factor, keeping the maximum value for every non-overlapping subregion of size $p_\text{row} \times p_\text{col}$ in the images.
\\ \textit{Fully Connected Layer (FC)}: this is the standard layer of a multi-layer network.
It performs a linear multiplication of the input vector by a weight
matrix. 

For a more detailed description and for the MPCNN back--propagation steps we refer the reader to the relevant literature~\cite{ciresan:2011b,masci:2012ijcnn}.

\noindent{\bf Image Segmentation with MPCNN}
Given a trained MPCNN, a straightforward approach for segmenting an unseen image requires to evaluate the net on every patch contained in it.  This results in many redundant computations, since different patches overlap.  A recent optimized approach \cite{giusti:TR0113} efficiently forward-propagates the whole image  (instead of a single patch) through the net, resulting in a teoretical  speed-up of almost three orders of magnitude during testing.

Here we extend this approach to speed up network training.
We define a novel neural network layer type called MaxPoolingFragment; 
 then, we derive the back--propagation procedure and show that the new model learns orders of magnitude faster than patch-based approaches.

\section{Method}
\label{sec:method}
\subsection{Notation}


The following notation is adopted. 
The set of training images is indicated by $\mathbf{X}$; the corresponding ground-truth annotations by $\mathbf{T}$ (thus mapping a class to each pixel); $x_i$ and $t_i$ refer to a particular training and testing image, respectively.
The net is parametrized by $\Theta$, the union of all parameters of all layers, and consists of a list of concatenated layers. The objective function of the minimization problem is denoted $L(\Theta; \mathbf{X}, \mathbf{T})$. 
A layer is a function mapping input storage to output storage.

Following earlier notation \cite{giusti:TR0113} we define such a storage as the set $\mathbf{F} = \{\cup_{i=1}^N f_i \}$, where each $f_i$ represents a stack of maps, here denoted as {\em Fragments}.  For example, the input layer will be a storage $\mathbf{F}^\text{input}$, with cardinality $1$.   $\mathbf{F}^\text{input}$ contains a single fragment, corresponding to the input image $x_i$.
This architecture strongly differs from conventional MPCNN by the choice of storage structure. In standard models every storage is a container of a stack of maps.  Instead, in our case the same data is necessarily split into a number of fragments.

\subsection{The MaxPoolingFragment (MPF) layer}

We can now introduce our MP layer extension, the
{\bf MaxPoolingFragment} (MPF).
Given an input image $x_i$, a conventional $k \times k$ MP layer
produces a 
smaller image, for which only a single value in each non-overlapping $k \times k$ neighborhood is kept.

When an MPCNN is applied with a sliding window though, forwarding every patch in the image causes redundant computations. While a convolutional layer can be applied directly to the entire input image to produce all results for all possible patches, an MP layer cannot.
The forward pass of our MPF layer closely follows the detailed description by Giusti et al.~\cite{giusti:TR0113}.
With a MPF layer there will be $k^2$ different offsets in the input map, each one producing an output fragment. Thus, if the number of fragments in input is $|\mathbf{F}^{\text{in}}|$, we will have $|\mathbf{F}^{\text{in}}|k^2$ fragments in total. All
 redundant computations are removed.

%
\begin{figure}[htbp]
\begin{center}
\includegraphics[width=1.\linewidth]{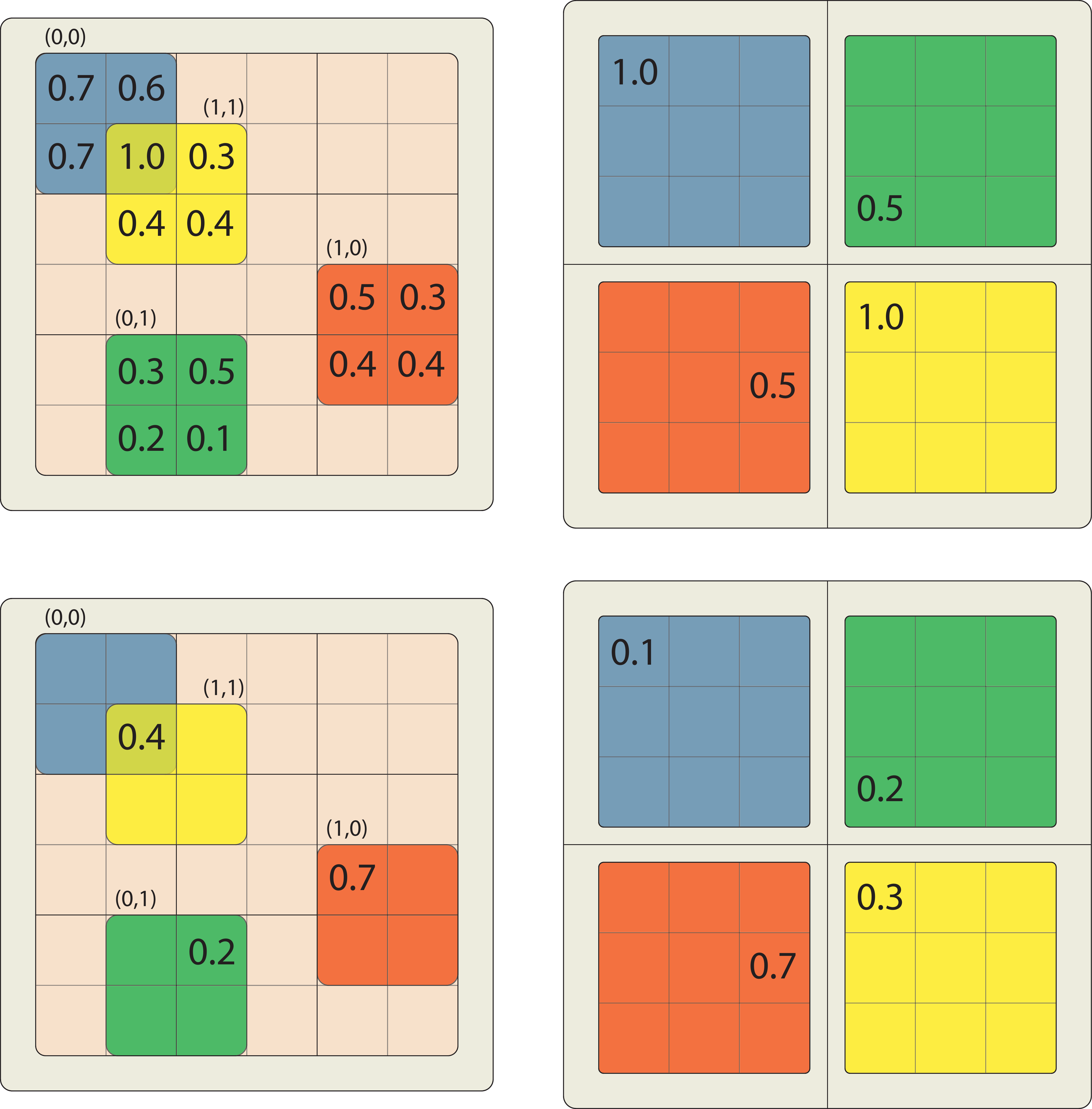}
\caption{MPF layer for the $2 \times 2$ pooling case. {\em Top}: Forward pass, {\em Fragments} (0,0) and (1,1) share the same maximal element; {\em Bottom} Back--propation pass where partial derivatives are pushed back to the previous layer in the hierarchy; the partial results of each {\em Fragment} are summed together. }
\label{fig:mpf}
\end{center}
\end{figure}

From a software engineering perspective,
an MPF layer can also be seen as a collection of MP layers, one for each generated fragment.

Consider a training image $x_i$ and its ground truth $t_i$. By means of the newly-defined MPF layer, we can quickly forward-propagate the network on the whole image to generate an output image $\hat{x_i}$, in which every pixel is replaced by the output of the corresponding MPCNN applied to every patch in $x_i$.
We can now compute the partial derivative of the loss function $L$ w.r.t. $\hat{x_i}$; e.g. $\hat{x_i} - t_i$ when minimizing the mean squared error loss with linear outputs. This is the first step of the back--propagation algorithm.

Now we are ready to present our main contribution, namely, a procedure which is able to process these errors without considering each output pixel separately.  This is made possible by an algorithm to compute the partial derivative of a MPF layer output w.r.t. its input, provided below.

\subsection{Back--propagation through an MPF layer}
As our framework uses fragments, we first have to redefine how a layer processes its input and computes its partial result of back--propagation.
This generalizes MPCNN operations as explained in object-oriented Matlab pseudocode by Algorithms \ref{alg:gen_layer_fwd} and \ref{alg:gen_layer_grad}.

\begin{algorithm}[ht]
 \SetAlgoLined
 \KwData{Layer $l$, Input storage $\mathbf{F}^{\text{in}}$}
 \KwResult{Output storage $\mathbf{F}^{\text{out}}$}
\For{i=1 \emph{\KwTo} $\mathbf{F}^{\text{in}}\rightarrow$nFragments()} {
	$\mathbf{F}^{\text{out}}$ = $\mathbf{F}^{\text{out}} \cup l\rightarrow$fwd($f^{\text{in}}_i$)\;
}
\caption{Pseudocode for the forward pass of a layer in our MPCNN framework operating on fragments. The backward pass is similarly derived. $\cup$ indicates the concatenation of two sets. An MPF produces a set of fragments.}
\label{alg:gen_layer_fwd}
\end{algorithm}
\begin{algorithm}[ht]
 \SetAlgoLined
 \KwData{Layer $l$, Input storage $\mathbf{F}^{\text{in}}$, Partial result of back--propagation $\mathbf{F}^{\delta}$}
 \KwResult{g: $\frac{\partial\mathrm{L(\Theta)}}{\partial l\rightarrow params}$}
\For{i=1 \emph{\KwTo} $\mathbf{F}^{\delta} \rightarrow$ nFragments()} {
	g = g + $l\rightarrow$grad($f^{\delta}_i$)\;
}
\caption{Gradient computation pseudocode of a generalized MPCNN layer used to operate in conjunction with a MPF layer. Gradients are accumulated as the same function is applied to every input fragment.}
\label{alg:gen_layer_grad}
\end{algorithm}

\begin{figure}[htbp]
\begin{center}
\includegraphics[width=1.\linewidth]{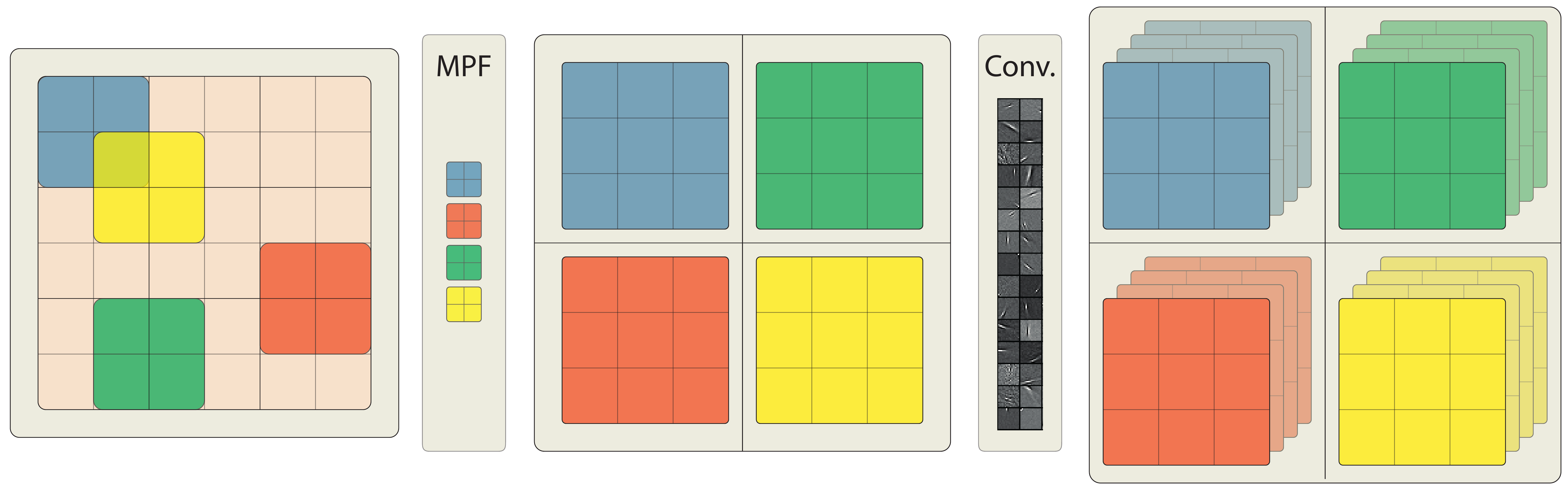}
\caption{Illustrative example of how subsequent convolutional layers of a MPF layer work. The same operation is applied to each output fragment; the gradient is  the sum of the gradients of the fragments.}
\label{fig:mpf_conv}
\end{center}
\end{figure}

This new neural network interface makes it much easier to derive and implement the backward pass for a MPF layer. Figure \ref{fig:mpf} gives an illustrative example of how a MPF layer works in the $2 \times 2$ pooling case. The output consists of $4$ fragments, each containing as many maps as the input layer, indexed accordingly. 
We also exemplify the case where the same element (pixel with value $1.0$ in Figure~\ref{fig:mpf}--top) is the 
maximum for two different offsets of the pooling kernel (respectively $(0, 0)$ and $(1, 1)$), generating the corresponding output fragments (respectively blue and yellow).
During back--propagation the partial results of differentiation, shown in Figure~\ref{fig:mpf}--bottom--right, are processed for every fragment and then summed together. 
The subsequent convolutional layer processes the $4$ fragments through the interface of Algorithm~\ref{alg:gen_layer_fwd}, as shown in Figure~\ref{fig:mpf_conv}, and computes the gradient by summing gradients of  4 fragments.
Pseudocode for both operations is shown in Algorithms~\ref{alg:mpf_fwd} and \ref{alg:mpf_bkp}.

%
\begin{algorithm}[ht]
 \SetAlgoLined
 \KwData{Input storage $\mathbf{F}^{\text{in}}$, Pooling kernel $P$}
 \KwResult{Output storage $\mathbf{F}^{\text{out}}$, MP indices $\mathbf{mpIDXS}$}
 $\mathbf{F}^{\text{out}} = \{\}$ \;
\For{i=1 \emph{\KwTo}$\mathbf{F}^{\text{in}}\rightarrow$nFragments()} {
\For{j=1 \emph{\KwTo}size(P, 1)*size(P, 2)} {
 	[r, c] = ind2sub(j, size(P)) \; 
	[a, b] = MP\_fwd($f^{\text{in}}_{i}$(r:end, c:end), P) \;
	$\mathbf{F}^{\text{out}}$ = $\mathbf{F}^{\text{out}} \cup a$ \;
	$\mathbf{mpIDXS}$ = $\mathbf{mpIDXS} \cup b$ \;
}
}
\caption{Forward pass of an MPF layer. This algorithm produces a set of fragments for each input fragment, one for each offset in the pooling kernel. The output is their union, as shown in Figure~\ref{fig:mpf}. MP($f^{\text{in}}$(r:end, c:end), P) indicates the usual MPCNN downsampling operation applied to the fragment $f^{\text{in}}$. Within the input maps, $\textbf{mpIDXS}$ stores the index of every maximum value produced by the pooling operation; it is used to back--propagate sub--gradients.}
\label{alg:mpf_fwd}
\end{algorithm}
\begin{algorithm}[ht]
 \SetAlgoLined
 \KwData{Input storage $\mathbf{F}^{\text{in}}$, Result of fwd $\mathbf{F}^{\text{out}}$, $P$, $\mathbf{mpIDXS}$}
 \KwResult{Output storage $\mathbf{F}^{\delta}$}
\For{i=1 \emph{\KwTo}$\mathbf{F}^{\text{out}} \rightarrow$nFragments()} {
\For{j=1 \emph{\KwTo}size(P, 1)*size(P, 2)} {
	s = findSourceFragment(i,j) \; 
	$f^{\delta}_{s}$ = $f^{\delta}_{s}$ + MP\_bkp($f^{\delta}_{s}$, $mpIDXS_{s}$) \;
}
}
\caption{Back--propagation pass of a MPF layer. The algorithm produces a set of fragments equal to the one of the input layer $\mathbf{F}^{\text{in}}$ used in Algorithm~\ref{alg:mpf_fwd}. MP\_bkp places the partial result of the chain rule of derivatives in the position indicated in $mpIDXS_{s}$. As each element might have contributed multiple times as shown in Figure~\ref{fig:mpf}--bottom, they need to be accumulated in the output layer $\mathbf{F}^{\delta}$.}
\label{alg:mpf_bkp}
\end{algorithm}
%




\section{Results}
\label{sec:results}
We validate the proposed approach on two different applications, namely membrane segmentation (Section~\ref{sec:membrane}) and steel defect detection (Section~\ref{sec:steel}).

In both applications, networks are trained to minimize the multi-class cross-entropy loss (MCCE) -- a commonly-used error function for classification tasks.  This error is computed by considering each pixel independently. Full connectivity is used for the convolutional layers. With C $7 \times 7 \times 8$ we indicate a layer with 8 output maps and filters of size $7 \times 7$. 


Our framework is implemented on CPU with Matlab and uses Intel Performance Primitives to perform convolutions through a Mex--function.

\subsection{Membrane Segmentation}
\label{sec:membrane}
We use the public dataset of the ISBI 2102 Electron Microscopy Segmentation challenge \cite{10.1371/journal.pbio.1000502}.
It consists of a volume of $30$ gray level images of size $512\times512$ pixels.

As in previous work \cite{Ciresan:2012f} we exploit the rotational invariance of the problem and synthesize additional training images by arbitrarily rotating and flipping the given training images.  Also, pixels outside of the boundary of testing images are synthesized by mirroring -- which allows to preserve the size of the output.
We consider the network architectures N3 and N4 of Ciresan et al.~\cite{Ciresan:2012f}, which contributed to the top-scoring entry in the challenge.

We train our system using stochastic gradient descent safe-guarded by the Armijo rule, updating the weights after every image has been presented to the net. Convergence is reached generally after 100 epochs, when the network has seen 3000 different images, the equivalent of roughly 390 million patches.

\begin{figure}[htbp]
\begin{center}
\includegraphics[width=1.\linewidth]{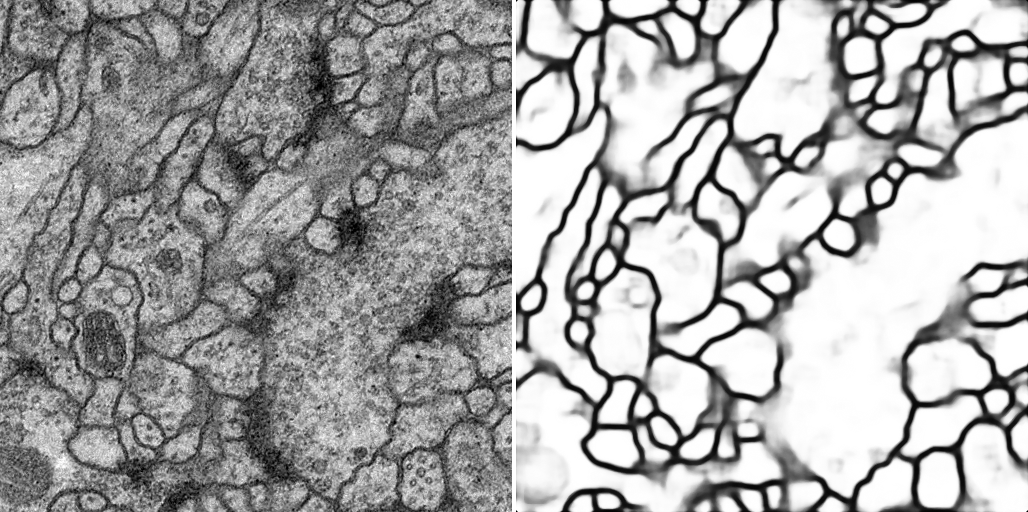}
\caption{A slice of the test set segmented using network N3 trained by our novel approach.}
\label{fig:membrane}
\end{center}
\end{figure}

%
%
Figure \ref{fig:membrane} shows a segmentation example for a test slice.
Table \ref{tab:membrane_times} compares our training times with the training times of the highly-optimized GPU patch-based approach~\cite{Ciresan:2012f}: although our implementation runs on the CPU in the Matlab environment, it yields a huge speed--up.
On the other hand, the segmentation performance of the resulting network exhibits negligible differences: 6.8\% vs 6.6\% pixel error rates for N3, respectively, for our method and the one of Ciresan et al.~\cite{Ciresan:2012f}. Errors are evaluated directly on the competition server.

%

\begin{table}[htdp]
\caption{Comparison of training times for the Membrane. The overhead for generating the transformed samples is also included in the overall computation. The relative speed--up of our method is shown in parenthesis.}
\begin{center}
\begin{tabular}{c c c}
	& \multicolumn{1}{c}{ Patch (GPU) \cite{Ciresan:2012f}} 	&  Image (CPU, Matlab), ours \\
 	 & patches/s & patches/s \\
\hline
\hline
 \multicolumn{1}{r}{ N3 } 	& 260  & 4500 (17$\times$ speed--up) \\
\multicolumn{1}{r}{ N4 }  & 130  & 3000 (23$\times$ speed--up) \\
\end{tabular}
\end{center}
\label{tab:membrane_times}
\end{table}

\subsection{Steel Defect Detection} 
\label{sec:steel}
\begin{figure}[htbp]
\begin{center}
\includegraphics[width=1.\linewidth]{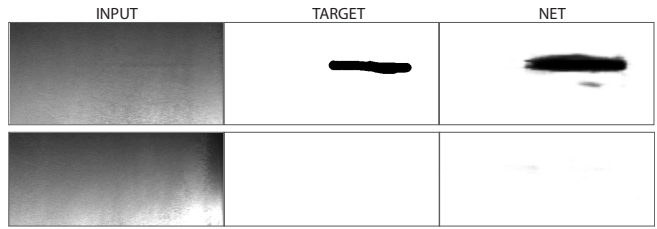}
\caption{A typical steel defect example. Segmentation is almost perfect, illustrating  the power of the proposed approach for industrial applications.}
\label{fig:steel}
\end{center}
\end{figure}

We use a proprietary dataset from ArcelorMittal, consisting of 534 images, each with resolution $550\times240$.
 Images are acquired with a matrix camera directly from a production plant.  Illumination is highly variable, and many images are severely under- or over-exposed, which hinders na\"{i}ve processing techniques.
70 of such samples contain a defect 
which covers part of each image. This type of defect is very difficult to detect due to its variable and subtle appearance (see Figure~\ref{fig:steel}).  A ground-truth segmentation of the area (if any) containing the defect of each image is given.
We use 50\% randomly--sampled images for training, 25\% for validation, and the rest for testing.

We consider the problem of segmenting the defect (if any) in each testing image.  Note that this is a harder problem than simply detecting whether there is a defect somewhere in the image -- which can be solved using a threshold on the number of pixels classified as defect.

The network operates on a $31\times31$ window
and has the following structure: C $7\times7\times8$, MPF $2\times2$, C $5\times5\times8$, MPF $2\times2$, C $5\times5\times8$, FC $100$, FC $2$.
We use LBFGS, which delivered the best performance. We also down--sample the images by a factor of 4 to further speed up learning.
We minimize the MCCE loss function per class because of the unbalanced dataset. There are in fact only very few pixels which correspond  to the defect, therefore learning is prone to na\"ive convergence to solutions which always favor predicting the background.

Every training epoch takes on average 44s 
(also accounting for the overhead due to LBFGS optimization). This amounts to 0.16s per image on a i7--2600 quad-core machine, where the whole system is trained in two hours.  Because an image contains roughly 8.2K patches, our system is effectively processing 50K patches per second.  Training on a patch-by-patch fashion is significantly slower even when highly optimized and implemented on GPU processors.

Segmenting a new image requires 0.07s, which makes online realtime implementation feasible for practical deployment within routine steel production activities.
In order to assess the segmentation performance of our model, we sample 5K positive and 5K negative pixels from the test set. This produces an unbiased evaluation to measure the per--pixel error rate. Random guessing reaches only 50\% error, while our MPCNN obtains 5.8\%.
Figure~\ref{fig:steel} shows a typical segmentation result for a test set image. 
%
%
%

We design our detection pipeline as follows.
After learning an MPCNN on the training set, we determine on the validation set the threshold yielding the best detection performance.

A given image is flagged as containing a defect if the number of ``defect'' pixels it contains exceeds a given threshold.  The threshold is set to $5000$ (i.e. half the area of the smallest conceivable area for a defect) and is not critical.

Table \ref{tab:detection_results} shows detection performance. Our method makes only 3 mistakes and correctly detects 3 additional defects mislabeled during annotation. 


%
\begin{table}[htdp]
\caption{Detection error results of our efficient learning framework for MPCNN. Test evaluation times for a given image are also reported along with the patch based evaluation with equal implementation (e.g. Matlab). }
\begin{center}
\begin{tabular}{c c c c}
	 &Test err \% & Patch (CPU) & Image (CPU) \\
\hline
\hline
 \multicolumn{1}{r}{  }  & 2.3  & 110s & 0.07s (1500x speed--up) \\
\end{tabular}
\end{center}
\label{tab:detection_results}
\end{table}

\section{Conclusions}
\label{sec:conclusion}
We introduced a fast and efficient algorithm to train MPCNN for image segmentation. The network is able to process the whole image at once without having to consider separate patches. This greatly speeds up training in comparison to approaches that process each patch independently -- including those optimized on GPU.  No significant loss in accuracy is observed.  Now we can train huge network architectures on large datasets within manageable timeframes.

We achieve state-of-the-art performance on the challenging membrane dataset used for the ISBI EM segmentation challenge. In an application to automatic steel inspection we reach realtime processing speed, achieving a speed--up factor of 1500 w.r.t. the corresponding patch-based method.


\bibliographystyle{IEEEbib}
\small
\bibliography{references}

\begin{thebibliography}{10}

\bibitem{boykov2001fast}
Y.~Boykov, O.~Veksler, and R.~Zabih,
\newblock ``Fast approximate energy minimization via graph cuts,''
\newblock {\em Pattern Analysis and Machine Intelligence, IEEE Transactions
  on}, vol. 23, no. 11, pp. 1222--1239, 2001.

\bibitem{tsai2003shape}
A.~Tsai, A.~Yezzi~Jr, W.~Wells, C.~Tempany, D.~Tucker, A.~Fan, W.E. Grimson,
  and A.~Willsky,
\newblock ``A shape-based approach to the segmentation of medical imagery using
  level sets,''
\newblock {\em Medical Imaging, IEEE Transactions on}, vol. 22, no. 2, pp.
  137--154, 2003.

\bibitem{Ciresan:2012f}
Dan~C. {Cire{\c s}an}, Alessandro Giusti, Luca~M. Gambardella, and J{\"u}rgen
  Schmidhuber,
\newblock ``Deep neural networks segment neuronal membranes in electron
  microscopy images,''
\newblock in {\em NIPS}, 2012.

\bibitem{farabet-pami-13}
Clement Farabet, Camille Couprie, Laurent Najman, and Yann LeCun,
\newblock ``Learning hierarchical features for scene labeling,''
\newblock {\em IEEE Transactions on Pattern Analysis and Machine Intelligence},
  2013,
\newblock in press.

\bibitem{Turaga:2010}
Srinivas~C. Turaga, Joseph~F. Murray, Viren Jain, Fabian Roth, Moritz
  Helmstaedter, Kevin Briggman, Winfried Denk, and H.~Sebastian Seung,
\newblock ``Convolutional networks can learn to generate affinity graphs for
  image segmentation,''
\newblock {\em Neural Comput.}, vol. 22, no. 2, pp. 511--538, Feb. 2010.

\bibitem{Turaga:2009}
Srinivas Turaga, Kevin Briggman, Moritz Helmstaedter, Winfried Denk, and
  Sebastian Seung,
\newblock ``{Maximin affinity learning of image segmentation},''
\newblock in {\em Advances in Neural Information Processing Systems 22},
  Y.~Bengio, D.~Schuurmans, J.~Lafferty, C.~K.~I. Williams, and A.~Culotta,
  Eds., pp. 1865--1873. 2009.

\bibitem{masci:2012ijcnn}
Jonathan Masci, Ueli {Meier}, Dan~C. {Cire{\c s}an}, Fricout Gabriel, and
  J{\"u}rgen {Schmidhuber},
\newblock ``Steel defect classification with max-pooling convolutional neural
  networks,''
\newblock in {\em International Joint Conference on Neural Networks}, 2012.

\bibitem{ciresan:2011b}
Dan~C. {Cire{\c s}an}, Ueli {Meier}, Jonathan {Masci}, and J{\"u}rgen
  {Schmidhuber},
\newblock ``Flexible, high performance convolutional neural networks for image
  classification,''
\newblock in {\em International Joint Conference on Artificial Intelligence
  (IJCAI2011)}, 2011.

\bibitem{ciresan:2011a}
Dan~C. {Cire{\c s}an}, Ueli {Meier}, Jonathan {Masci}, and J{\"u}rgen
  {Schmidhuber},
\newblock ``A committee of neural networks for traffic sign classification,''
\newblock in {\em International Joint Conference on Neural Networks
  (IJCNN2011)}, 2011.

\bibitem{ciresan:2011c}
Dan~C. {Cire{\c s}an}, Ueli Meier, Luca~M. Gambardella, and J{\"u}rgen
  Schmidhuber,
\newblock ``Convolutional neural network committees for handwritten character
  classification,''
\newblock in {\em ICDAR}, 2011, pp. 1250--1254.

\bibitem{giusti:TR0113}
Alessandro Giusti, Dan~C. {Cire{\c s}an}, Jonathan Masci, Luca~M. Gambardella,
  and J{\"u}rgen Schmidhuber,
\newblock ``Fast scanning with deep neural networks,''
\newblock Tech. {R}ep. IDSIA-01-13, Istituto Dalle Molle di Studi
  sull'Intelligenza Artificiale (IDSIA), 2013.

\bibitem{10.1371/journal.pbio.1000502}
Albert Cardona, Stephan Saalfeld, Stephan Preibisch, Benjamin Schmid, Anchi
  Cheng, Jim Pulokas, Pavel Tomancak, and Volker Hartenstein,
\newblock ``An integrated micro- and macroarchitectural analysis of the
  drosophila brain by computer-assisted serial section electron microscopy,''
\newblock {\em PLoS Biol}, vol. 8, no. 10, pp. e1000502, 10 2010.

\end{thebibliography}

\end{document}